\documentclass[letterpaper]{article} 
\usepackage{xai4sci}
\usepackage{times}  
\usepackage{helvet}  
\usepackage{courier}  
\usepackage[hyphens]{url}  
\usepackage{graphicx} 
\usepackage{natbib}  
\usepackage{caption} 
\usepackage{algorithm}
\usepackage{algorithmic}

\usepackage{xcolor}

\usepackage{amsmath}

\usepackage{newfloat}
\usepackage{listings}
\DeclareCaptionStyle{ruled}{labelfont=normalfont,labelsep=colon,strut=off} 
\floatstyle{ruled}
\newfloat{listing}{tb}{lst}{}
\floatname{listing}{Listing}
\title{Studying the Impact of Latent Representations in Implicit Neural Networks for Scientific Continuous Field Reconstruction}
\author{
    Wei Xu\textsuperscript{\rm 1},
    Derek Freeman DeSantis\textsuperscript{\rm 2},
    Xihaier Luo\textsuperscript{\rm 1},
    Avish Parmar\textsuperscript{\rm 1},\\
    Klaus Tan\textsuperscript{\rm 3},
    Balu Nadiga\textsuperscript{\rm 2},
    Yihui Ren\textsuperscript{\rm 1},
    Shinjae Yoo\textsuperscript{\rm 1}
}
\affiliations{
    \textsuperscript{\rm 1}Brookhaven National Laboratory\\
    \textsuperscript{\rm 2}Los Alamos National Laboratory\\
    \textsuperscript{\rm 3}Not Affiliated\\

    xuw@bnl.gov, ddesantis@lanl.gov, xluo@bnl.gov, aparmar@bnl.gov, halfanteater87@gmail.com, balu@lanl.gov, yren@bnl.gov, sjyoo@bnl.gov
}

\usepackage{bibentry}

\begin{document}

\maketitle

\begin{abstract}
Learning a continuous and reliable representation of physical fields from sparse sampling is challenging and it affects diverse scientific disciplines. In a recent work, we present a novel model called MMGN (Multiplicative and Modulated Gabor Network) with implicit neural networks. In this work, we design additional studies leveraging explainability methods to complement the previous experiments and further enhance the understanding of latent representations generated by the model. The adopted methods are general enough to be leveraged for any latent space inspection. Preliminary results demonstrate the contextual information incorporated in the latent representations and their impact on the model performance. As a work in progress, we will continue to verify our findings and develop novel explainability approaches.
\end{abstract}

\section{Introduction}

Field reconstruction is the process involving the recreation of a physical field denoted as $\boldsymbol{u}$, depicting the spatial and temporal distribution of a specific parameter (e.g., temperature, velocity, or displacement). Its significance spans diverse scientific disciplines such as geophysics, climate science, and fluid mechanics, due to several key functions. First, it facilitates the extrapolation of values at unmeasured locations by leveraging available data~\citep{kutz2016dynamic}. It also contributes to the identification of patterns, trends, and variations within the investigated parameters~\citep{shu2023physics,wang2022detail}. Lastly, field reconstruction plays a crucial role in optimizing sensor placement and refining data collection strategies, thereby enhancing the effectiveness of experimental designs~\citep{fukami2021global,krause2008near}.

Recent progress in machine learning has sparked a growing interest in addressing field reconstruction challenges through implicit neural representations~\citep{xie2022neural,mildenhall2021nerf}. These representations harness multilayer perceptrons (MLPs) to represent images, videos, and 3D objects. Specifically, coordinate-based MLPs operate by taking low-dimensional coordinates, such as spatial position $(\boldsymbol{x})$ or time index $(t)$, as input and predicting the value of a learned signal $(\boldsymbol{u})$ for each coordinate point. The coordinate-based MLP effectively learns an implicit, continuous, and differentiable function, mapping input coordinates to the target signal $\boldsymbol{u} = F_{\theta} (\boldsymbol{x}, t)$.

Driven by the quest for an enhanced representation method, we introduce a context-aware indexing mechanism. Specifically, we integrate existing contextual information into the temporal coordinate $(t)$. Given the dynamic changes in the quantity and locations of available measurements over time $t$, we propose a framework where an encoder extracts a latent representation from real-time measurements. This latent representation is then utilized to steer the model towards the desired time instance. Paired with an implicit neural representation-based decoder, our proposed method excels in the seamless reconstruction of continuous fields~\cite{luo2024continuous}.

Meanwhile, as neural networks become more complicated, it is a grand challenge to increase the overall interpretability and trust especially when adopted by scientific users. For our field reconstruction model, the learned latent representation is an essential component and incorporates information about the measurements beyond temporal indexing. In this work, we present the application of several dimension reduction approaches and use them as explainability methods for latent representation understanding. Concretely, we leverage explainability tools to demonstrate that the latent representation has sufficiently encoded the real-time measurements. We take the use case of continuous field reconstruction using implicit neural networks for climate science and report the findings as a work in progress. 

\section{Related Work}
Explainable Artificial Intelligence (XAI) technique has flourished in the past decade. For the latent space understanding, since the latent vectors are difficult to interpret by humans directly, many methods are proposed to tackle the dimensionality challenges. Embedding techniques e.g., t-Distributed Stochastic Neighbor Embedding (t-SNE)~\cite{7b54165e73a3424b8820136bcf61ca89} are used to project high-dimensional data into 2D space and study whether the internal correlations of the data is preserved in the latent space. Clustering is another common method to apply to the latent vectors and demonstrate the relationship. Moreover, both linear and nonlinear methods are used to estimate the similarity between latent representations~\cite{article}. Developing visual analytics frameworks is another branch of research to facilitate user-directed analysis exploring the latent space interactively~\cite{SHEN202099}. We find that existing works for dimensionality and decomposition studies can be beneficiary for general-purpose latent understanding, which are less commonly adopted beyond t-SNE visualization and ablation studies.

\section{Methodology}

\subsection{The MMGN model}

Traditional implicit neural networks incorporate the time index along with spatial coordinates. 
Here, we present a neural network reconstruction method called MMGN (Multiplicative and Modulated Gabor Network)~\citep{luo2024continuous}, characterized by an encoder-decoder architecture. For the encoder, we propose to utilize measurements, represented as $u^1_t$, $u^2_t$, $u^3_t$,... at time $t$, for implicit model guidance instead of relying on $t$ for explicit pointing. Therefore, we construct an encoder $E(\cdot)$ to convert the observed measurements into a latent code $\boldsymbol{z}_t$. Subsequently, we employ a decoder $D(\cdot)$ that utilizes both the encoded information and spatial coordinates $\boldsymbol{x}$ to reconstruct the underlying physical field.

\begin{equation}
\begin{aligned}
\text{Baseline}: & u(\boldsymbol{x}, t) = F(\boldsymbol{x}, t) \\
\text{Proposed}: & u(\boldsymbol{x}, t) = D(\boldsymbol{x}, \boldsymbol{z}_t) = D(\boldsymbol{x}, E(u^1_t, u^2_t, \dots))
\end{aligned}
\end{equation}

\noindent \textbullet $\,$ \textbf{Encoder}: Autoencoders (AE) and their probabilistic variant, variational autoencoders (VAE)~\citep{kingma2014autoencoding}, are frequently employed for representation learning, leveraging their inherent latent variable structures. However, conventional AE and VAE encounter challenges when dealing with the variability of sensor locations and their fluctuations over time. While their graph counterparts can manage spatial variability, and modified versions address dynamic graphs, they become computationally demanding for extensive graphs and encounter difficulties with long-range dependencies. In contrast, auto-decoder, as exemplified in~\citet{park2019deepsdf}, demonstrates reduced underfitting and heightened flexibility. It accommodates observation grids of varying forms, including irregular or those existing on a manifold, without necessitating a specialized encoder architecture, provided the decoder shares the same property. Therefore, the auto-decoder is chosen as the backbone for the encoder due to its ability to accommodate free-formed observations, considering that the number and positions of available observations vary over time.

\noindent \textbullet $\,$ \textbf{Decoder}: The decoder inputs comprise two components: spatial coordinates ($\boldsymbol{x}$) and latent codes ($\boldsymbol{z}$). Initially, processing $\boldsymbol{x}$ through fully-connected feed-forward layers results in a coordinate-based MLP. Although such an MLP based on coordinates can provide a continuous representation, it encounters challenges in learning high-frequency signals, a phenomenon referred to as spectral bias. Recent studies suggest that addressing this issue involves incorporating positional encoding with Fourier features~\citep{tancik2020fourier} or introducing periodic non-linearities in the first hidden layer~\citep{sitzmann2020implicit}. In our approach, we employ Gabor filters $g(\cdot)$ over Fourier bases to transform the coordinates. Following the transformation of coordinates $\boldsymbol{x}$, we introduce a modulation step where the transformed coordinates $g(\boldsymbol{x})$ undergo modulation through a multiplicative layer, facilitating the integration of $g(\boldsymbol{x})$ and $\boldsymbol{z}$. Specifically, the decoder incorporates a multiplicative filter network (MFN)~\citep{fathony2020multiplicative} as its backbone network. This choice is motivated by the recursive mechanisms embedded in MFN, enabling improved fusion of spatial coordinates ($\boldsymbol{x}$) and latent codes ($\boldsymbol{z}$).

\noindent \textbullet $\,$ \textbf{Trained models}: The MMGN model used in this work is trained with 5\% sampling rate of the original dataset generated by the CESM2 climate model~\citep{danabasoglu2020community} that simulates Earth's climate states. We take monthly averaged global surface temperature data for model testing. That said, the network can be trained using a sampling rate at 5\% of the training data to recover a continuous represenation of it without substantial errors.  Moreover, by altering the hyperparameter $k$ defining the size of latent codes $\boldsymbol{z}_k$, we can train different MMGN models. In our previous study, we train $10$ models with latent sizes, ranging from $1$ to $512$ by doubling the latent size at a time, and collect the corresponding learned latent codes for all $T$ time steps. We found that increasing the latent size can reduce the reconstruction error and the improvement becomes more subtle when the latent size is moderate.

\subsection{Explainability approaches}
We adopt several explainability methods to conduct additional studies that further understand the impact of latent codes on the model performance and illustrate the contextual information latent codes integrate. \vspace{0.5em}

\noindent \textbullet $\,$ \textbf{Embedding and Clustering}
Latent representations are spatiotemporal vectors in high-dimensional space. To compare the representations in different latent spaces, we first adopt embedding techniques to project data into a 2D plane. For the demonstration, we use t-SNE to generate the distribution of $T$ latent vectors of each latent space and use K-means clustering to partition them into groups. Then, we visualize the distributions and quantify the statistics of clusters for all the latent spaces so that the variations can be observed. This helps to explain the impact of latent size on the acquired contextual information by the encoder $E(\cdot)$.

\noindent \textbullet $\,$ \textbf{Correlation Analysis}
Latent codes can also be represented by a $T \times k$ matrix where $k$-dimensional latent vectors are stacked in raster order. Comparing the latent spaces becomes the comparison of the corresponding matrices so that the rows represent samples and the columns represent features. In that sense, we can use correlation analysis to generate patterns of different latent spaces and compare their correlations. We leverage both Principal Component Analysis (PCA)~\cite{da6385d2-9c65-3860-bbcd-b821fdff69ff} and Canonical Correlation Analysis (CCA)~\cite{Härdle2007} in this study. 

\begin{figure*}[t]
\centering
\begin{tabular}{c c c}
\hline 
\vspace{-0.27cm} \\
\includegraphics[scale=0.18]{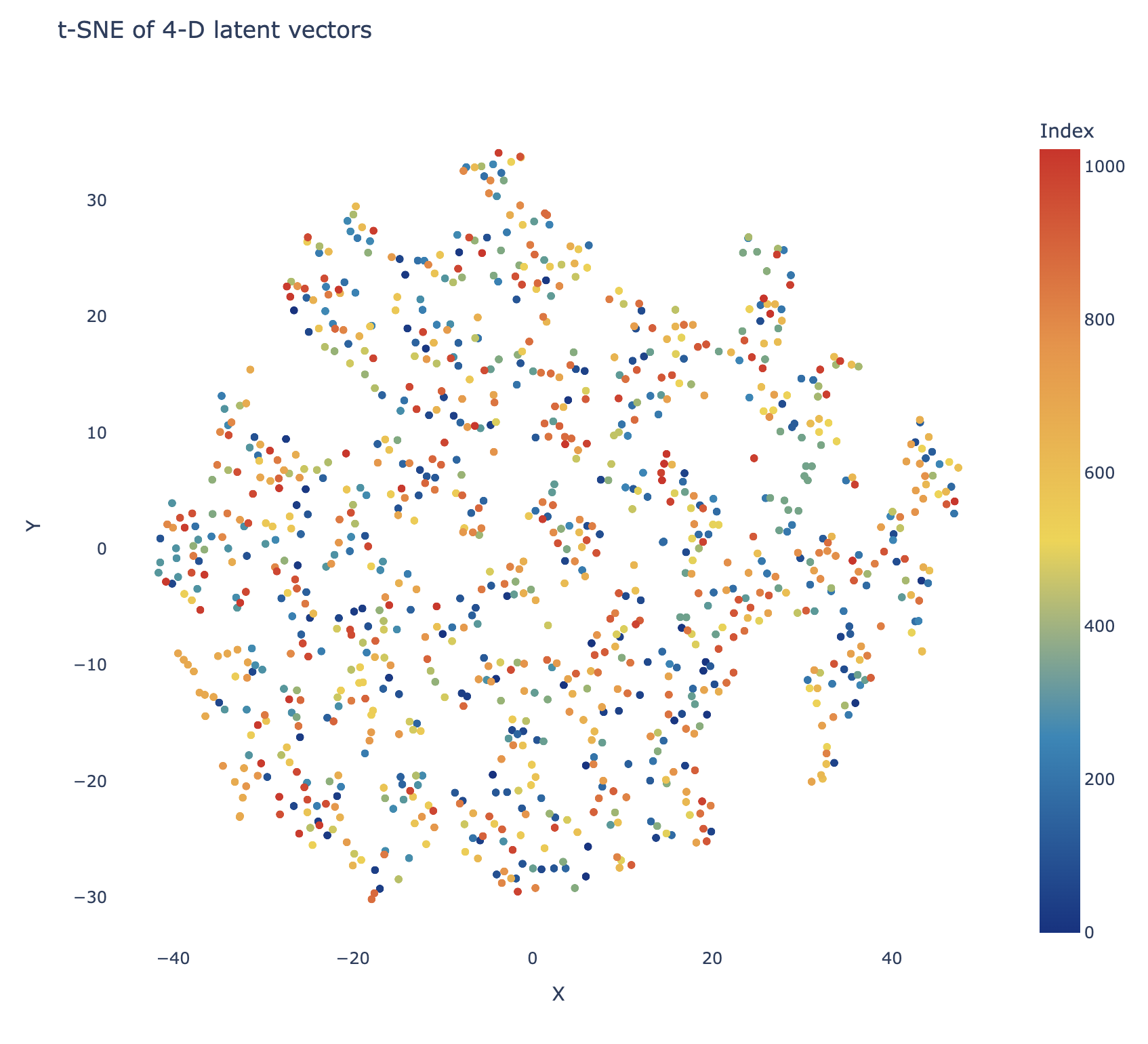} &
\includegraphics[scale=0.18]{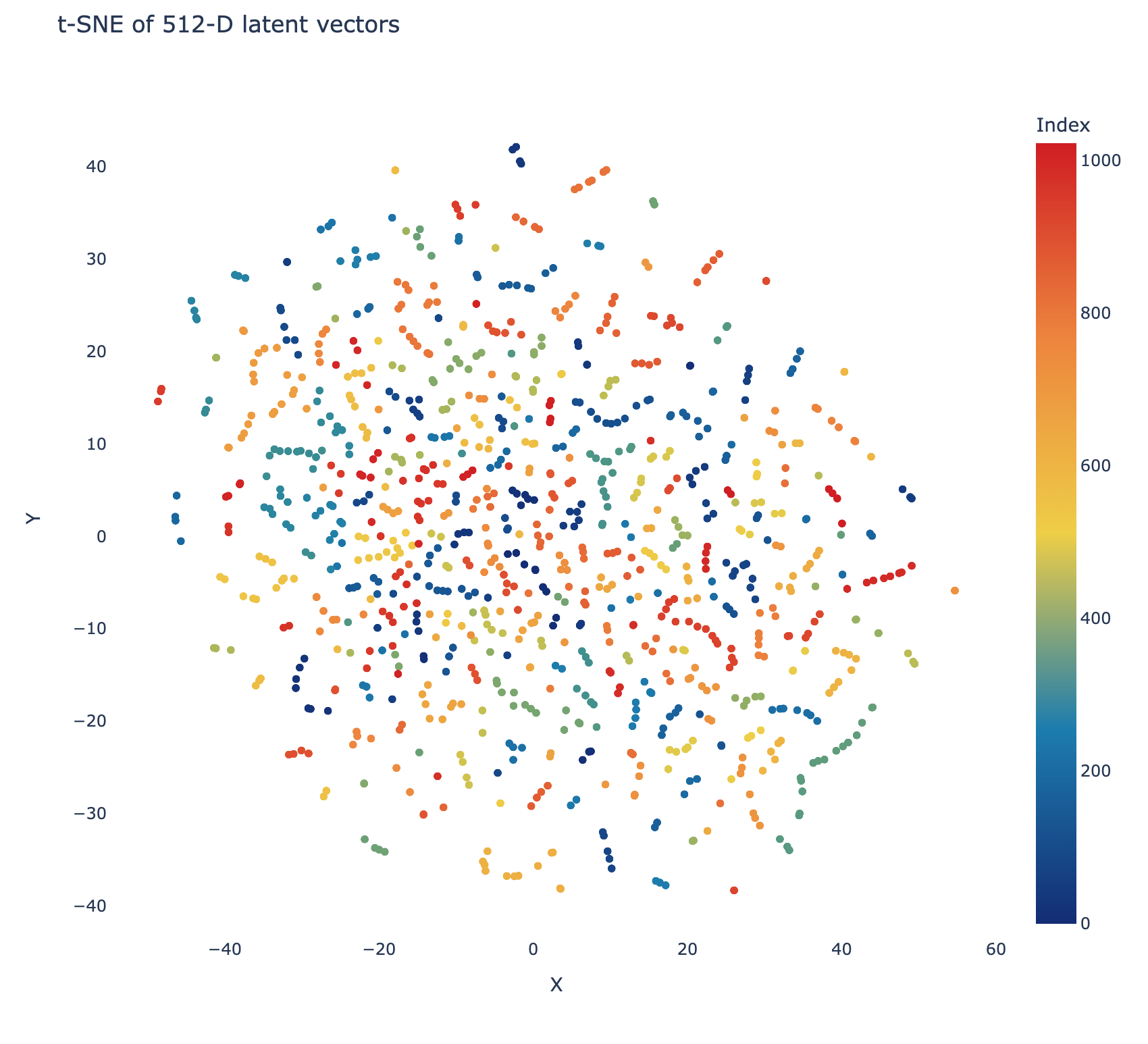} &
\includegraphics[scale=0.18]{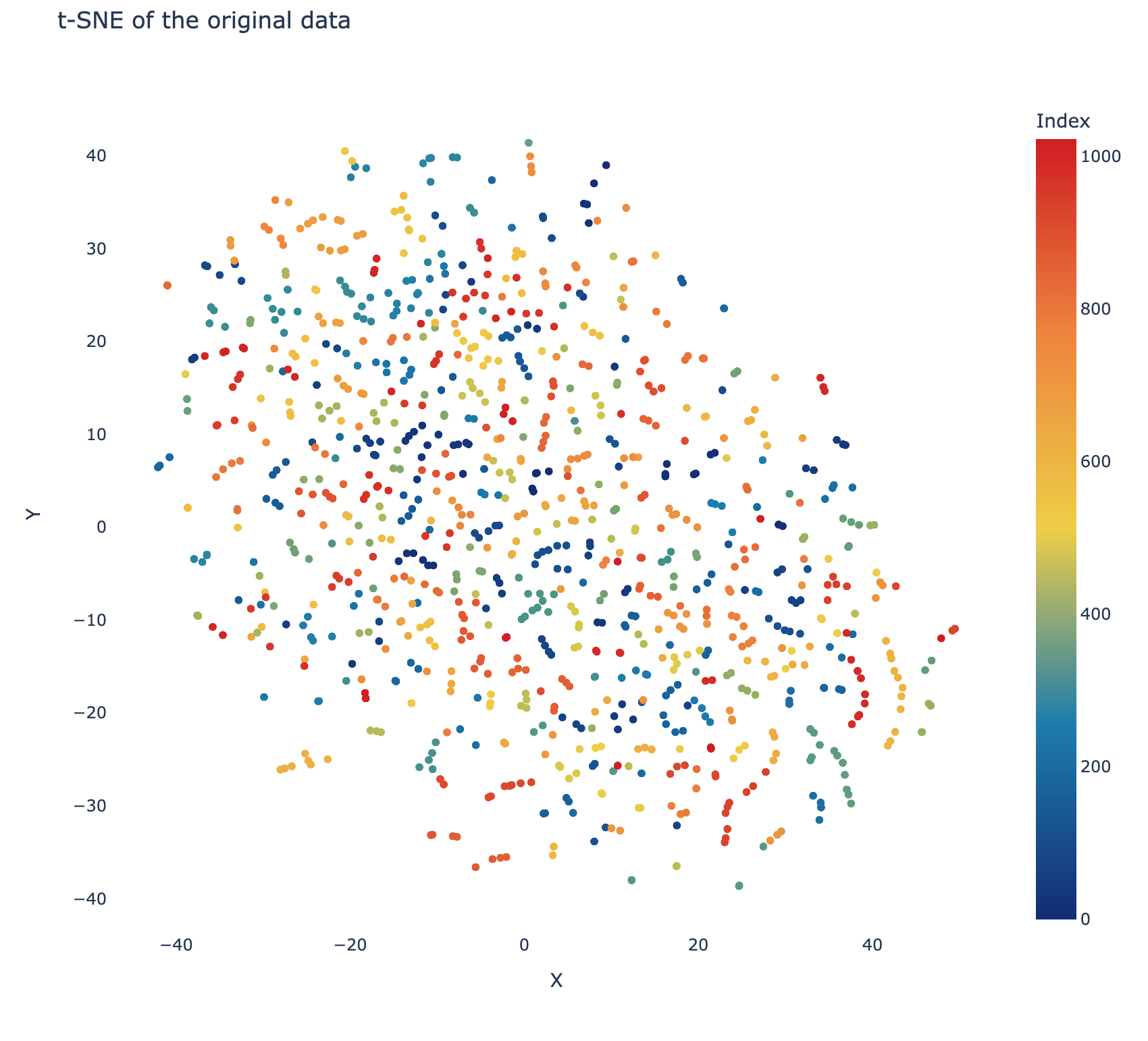} \\
\hline 
\vspace{-0.27cm} \\
\end{tabular}
\caption{The t-SNE distribution changes in three latent spaces: (left) 4-D, (middle) 512-D, and (right) the original data; the color represents the temporal indexing.}
\label{fig: tsne}
\end{figure*}

PCA identifies the direction i.e., the principal component, formed by a linear combination of original features that captures the maximal variance of the data. The explained variance ratio (i.e., normalized eigenvalue) is a measurement used to explain the proportion of variance of the matrix explained by the corresponding principal component. Thus, we generate $k$ principal components with their corresponding explained variance ratios for the matrix with $k$-dimensional latents and compare the trends. Meanwhile, we also apply PCA to the original data and generate $512$ explained variance ratios. By visualizing the trends of ratios, we could evaluate the coherence of the rank of information preserved in latent spaces and the original data. 

CCA provides a more direct way to identify and measure the associations between two sets of features, here two matrices. It determines pairs of canonical variates $u$ from the first set and $v$ from the second set so that they are maximally correlated with each other. The difference between PCA and CCA roots in that while PCA finds the linear combination of features that maximizes variance in the data, CCA looks for a linear combination of features in both data that have the strongest correlation between each other.

\noindent \textbullet $\,$ \textbf{Tensor Factorizations}
Tensor factorizations provide a method for effectively explaining surrogate model complexity. Tensor factorizations are techniques for decomposing multidimensional arrays (tensors) into simpler, constituent components. Among various tensor factorization techniques, Tucker Decomposition plays a pivotal role in analyzing and interpreting multiway data. This method decomposes a tensor into a core tensor multiplied by factor matrices along each dimension of the tensor.  The factor matrices describe the dominant modes along each variable, while the core tensor describes their interactions or mixing. Both the data used, and output derived from field reconstruction,  intrinsically arises as a tensor.  The Tucker decomposition is a natural tool to reveal the dominant modes and interactions within these data sets. See \cite{kolda2009tensor} for more details.



We leverage the Tucker Decomposition to assess how well MMGN represents the dominant modes (i.e., $\boldsymbol{x}$ and $t$) and their mixing. By taking the latent size at 256 for this study, the Tucker decomposition of both the training set and the model output are computed. The factor matrices and the mixing tensor are then compared to see how well MMGN does at reproducing the modes and mixing complexity of the Earth system data.

To compare the modes, we compute the Pearson correlation coefficient between the modes for each variable.  A high matching across modes in a variable indicates that MMGN has learned the data generating process for that particular variable.  To compare the cores, we compute the Tucker decomposition for cores shaped $[r,r,r]$ for each $r =1,2,\dots 200$.  The values of the core are then normalized, and the entropy of the core is computed. The entropy measures the complexity of the mixing at these different multiranks.  One should expect that the entropy of the MMGN model to be less than the entropy of the training data for each $r$, with exact matching indicating it has captured the mixing process between the variables.  

\noindent \textbullet $\,$ \textbf{Ablation Study}
Finally, an ablation study is designed to complement the previous effort inspecting the contribution of each dimension of the latent vector. By ablating one dimension at a time during the inference, we observed a reduced performance but the mean squared error (MSE) keeps mitigating when a bigger latent size is adopted. In this study, we want to decompose the overall error into the temporal and spatial counterparts and check if any resonance exists. 

\begin{figure*}[t]
\centering
\begin{tabular}{c c c}
\vspace{-0.27cm} \\
\includegraphics[scale=0.26]{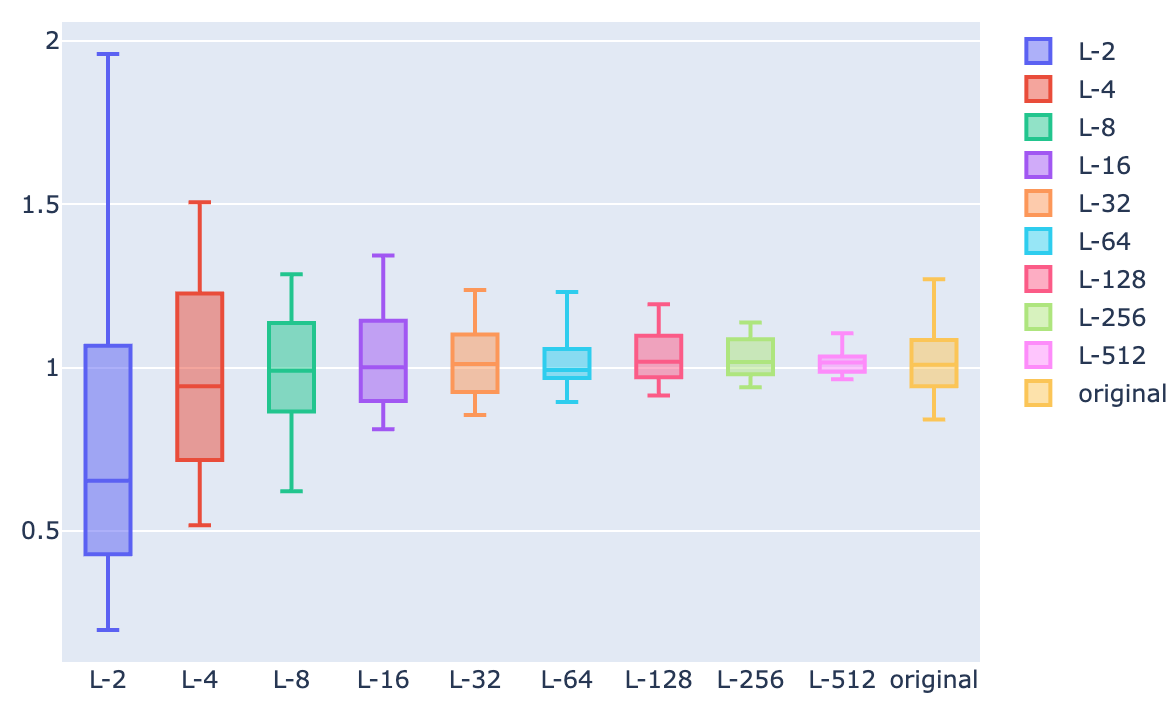} &
\includegraphics[scale=0.26]{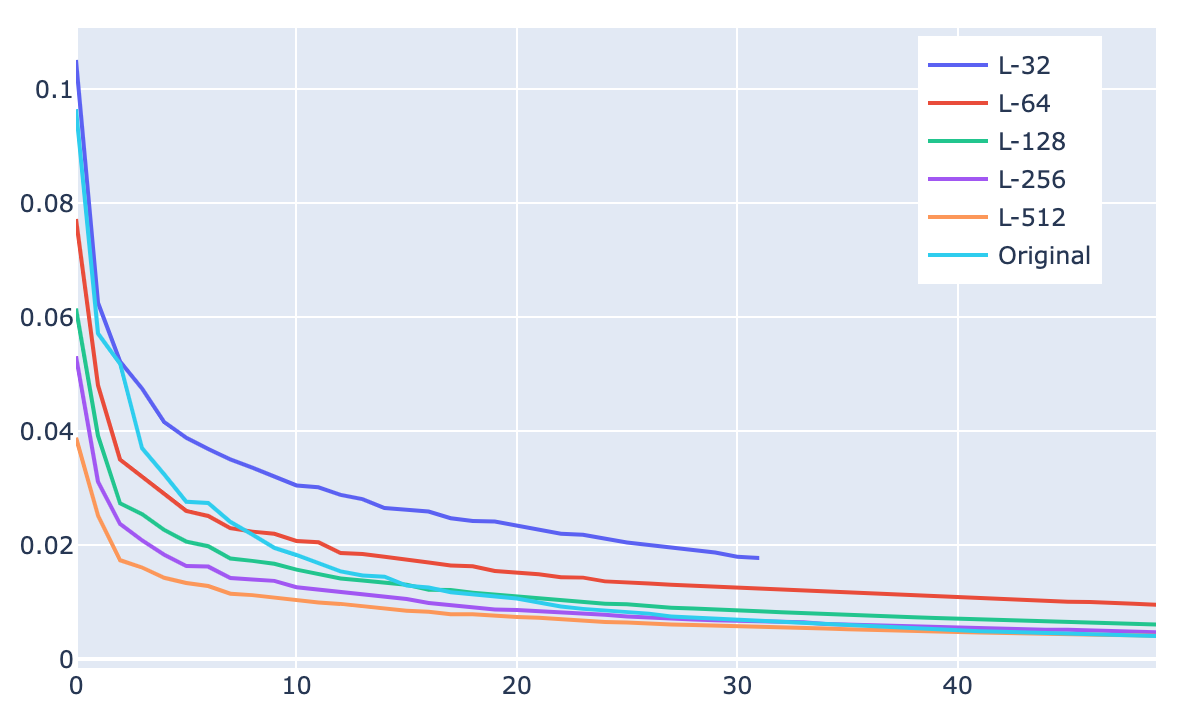} &
\includegraphics[scale=0.26]{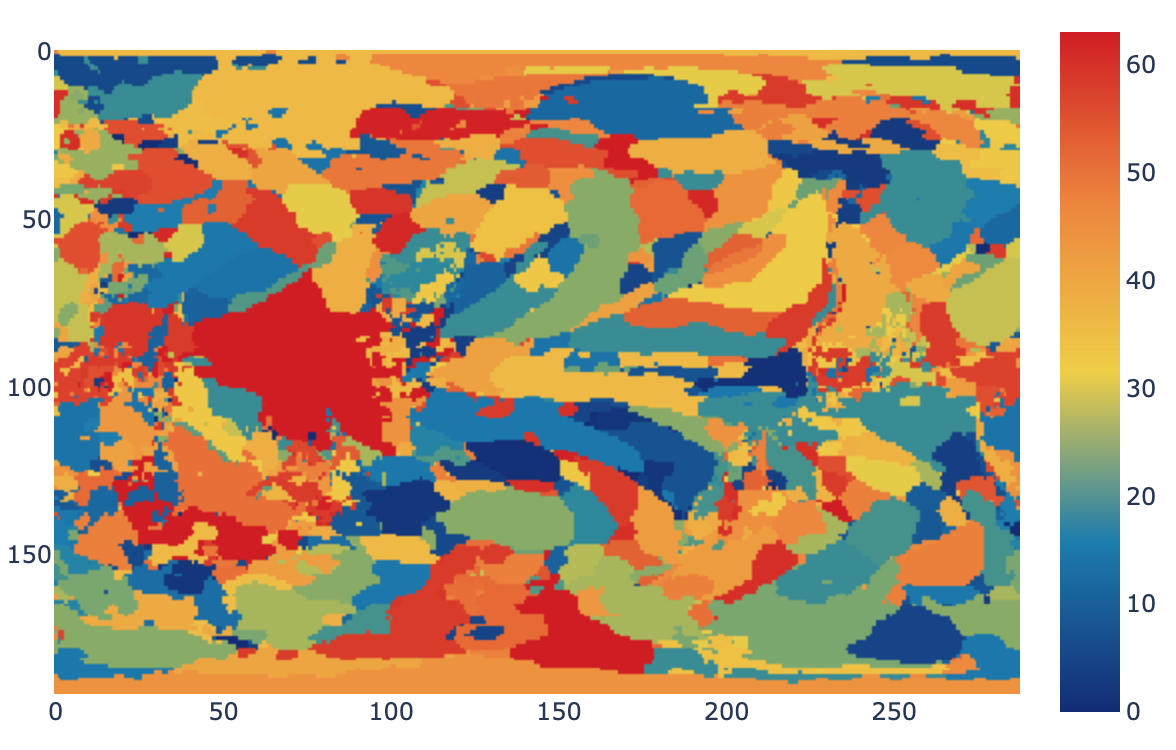} \\
\vspace{-0.27cm} \\
\end{tabular}
\caption{Left: The standard deviations of clusters diminish across latent spaces and they are compared with the ones of the original data (the rightmost boxplot). Middle: The trendings of explained variance ratios of the principal components for all latent spaces (illustrating only the top 5 sizes) and compare them with the trending of the original data. Right: Ablation result indicating spatial linkage of temporal indexing.}
\label{fig: comp}
\end{figure*}

\section{Results}

\begin{figure}
\centering
\includegraphics[scale=0.20]{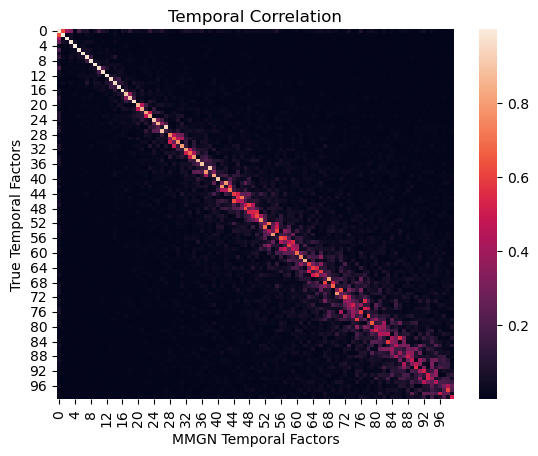}
\includegraphics[scale=0.20]{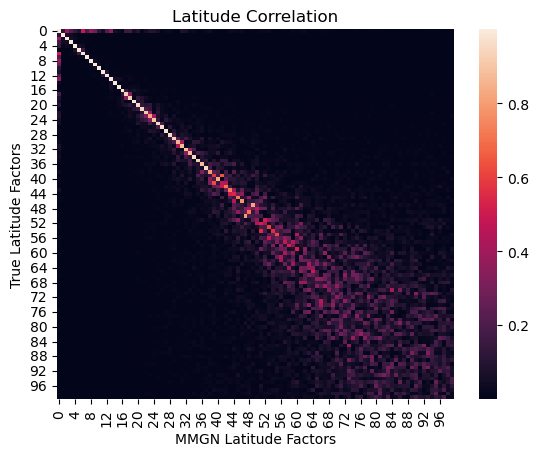}
\includegraphics[scale=0.20]{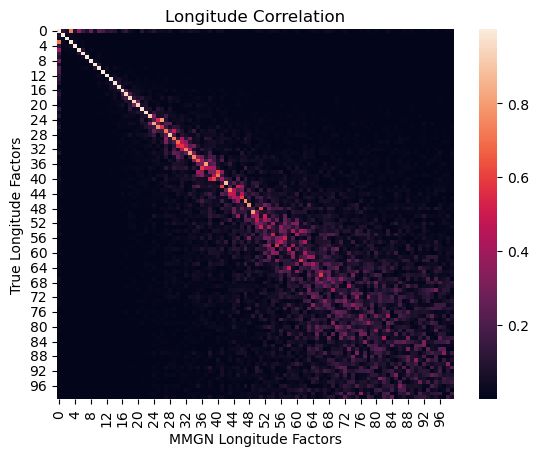}\\
\includegraphics[scale=0.26]{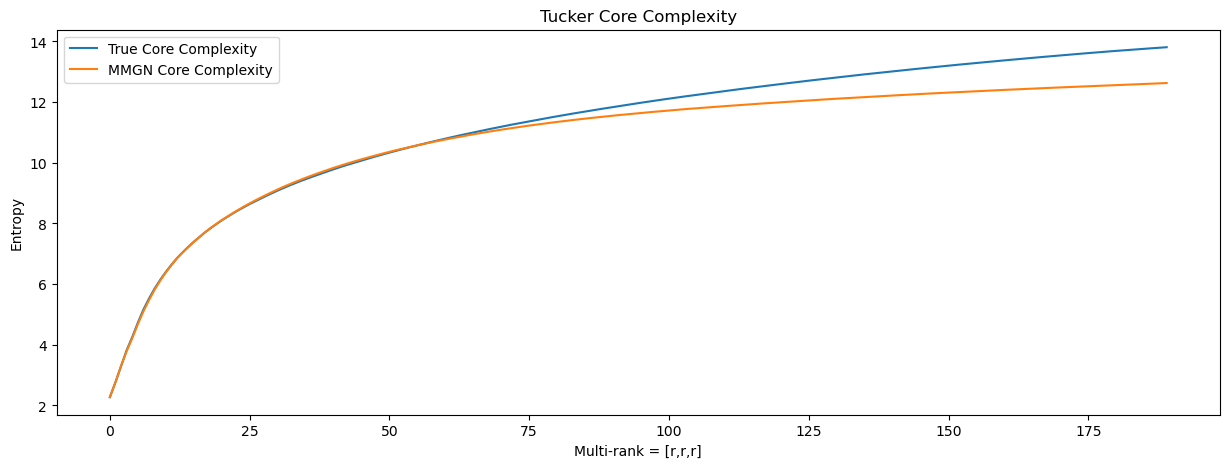}
\caption{Top: Correlation plots for the three variables.  Bottom: Complexity of Tucker core for different multi-ranks. }
\label{fig: TF}
\end{figure}

\noindent \textbullet $\,$ \textbf{Embedding and Clustering} The t-SNE visualizations are shown in Figure~\ref{fig: tsne} to illustrate how the distributions of latents in different latent spaces vary. Only the 4-D, 512-D, and the original spaces are shown for parsimony. Our goal is to understand the impact of the latent size on the learned representations. We observe that the lower dimensional space has more diverse but compact clusters and it gets more evenly spreading out when the dimension increases. We also find more string-shaped local clusters having the same color in the higher dimensional space. Since we use the temporal indexing $t$ as colors, these clusters are formed by the temporally adjacent latent vectors. The same pattern and distribution are observed in the original data. We conclude that the higher latent space better captures the global distribution of the original data, and to some extent, also maintains local coherence (string-shaped small clusters). Note here the use of t-SNE projections of 2-D latents instead of their original coordinates is meant for a fair comparison with other latent spaces. Then, the clustering results are generated and we collect the standard deviations $\sigma_1, ..., \sigma_n$ of the vectors in the corresponding $n$ clusters for all the latent spaces. As a preliminary result, the increased $\sigma$ pattern also confirms that the distribution shifts from compact and diverse groups to a wider space spanning as shown in Figure~\ref{fig: comp}(left). 

\noindent \textbullet $\,$ \textbf{Correlation Analysis} We aim to use correlation analysis to understand the impact of latent size in terms of rank complexity and potential information loss. First, the PCA results are shown in Figure~\ref{fig: comp}(middle) where the explained variance ratios of higher latent spaces are compared with the one of the original data. The slope of the curve suggests the rank of data complexity. For example, a steep slope means with the first few principal components the most variances can be preserved. Compared to the slope of the original data, we observe a similar slope starting from a latent size at 64 and beyond. Furthermore, we also generate the CCA canonical variate pairs between two cases: 1) two latent spaces and 2) a latent space and the original data. Then we use Pearson correlation coefficient to gauge their relations. We find that all the paired canonical variates are fully correlated for both cases. We conclude that any latent code is a reduced representation presenting consistent information about the original data. 

\begin{figure}[h]
\centering
\includegraphics[scale=0.4]{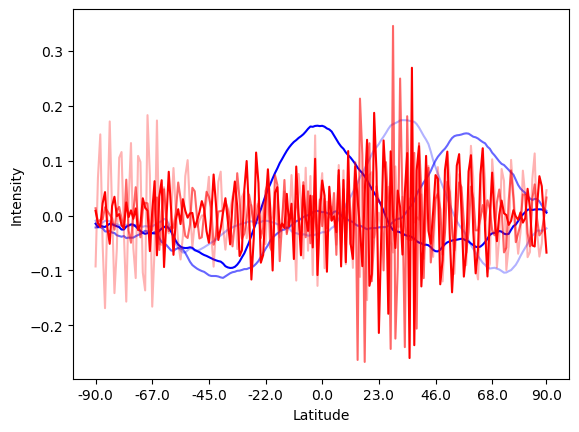}
\caption{Latitude modes $1,2,3$ in blue versus latitude modes $101,102,103$ in red. Higher order modes capture higher frequency information compared to lower order modes. }%
\label{fig: modes}
\end{figure}

\noindent \textbullet $\,$ \textbf{Tensor Factorizations} Similar to the purpose of adopting correlation analysis, the results of the tensor based analysis are shown in Figure \ref{fig: TF}.  For the factor matrices, we see that the MMGN model has strong agreement with the true factors for the first $20$ temporal and longitudinal components, and first $40$ latitudinal components.  These top modes are responsible for the majority of the reconstruction of the tensor (roughly $\sim 4 \%$ relative error), so agreement here indicates MMGN has learned the dominant spatial temporal patterns of the dataset.  The divergence of higher order modes suggests that MMGN fails to capture the higher frequency spatial components of the data (see Figure~\ref{fig: modes}). The fact that there is stronger agreement in the latitude modes indicates the model had an easier time capturing this component. 

From Figure~\ref{fig: TF} we see that the core of the MMGN model has complexity that matches the true data up to multi-rank $[75,75,75]$.  At this high of a rank, the tensor decomposition for both the true data and the MMGN output have a relative reconstruction error of $\sim 1 \%$. The agreement of model complexity up to this large multi-rank indicates that MMGN accurately captures the mixing processes of the underlying physical phenomenon. 

\noindent \textbullet $\,$ \textbf{Ablation Study} We collect the sets of temporal and spatial errors for the model of a latent size at 64. A preliminary result for the spatial linkage of the latent dimension is shown in Figure~\ref{fig: comp}(right). We use colors to indicate the latent dimension index (ranging between 0 and 63) so that once ablated it causes the most performance reduction at a spatial location. It is interesting to observe that individual latent dimensions contribute to adjacent regions for low-fidelity areas. This plot also shows the MMGN model can capture long-range dependencies. We will further verify this finding in future work.


\section{Conclusion}

The present studies leveraged dimensionality reduction approaches to explain the effect of the encoder latent space dimension has on the representation. These tools have  improved understanding of the learned latent representations. As a work in progress, we will keep verifying the current findings and work on potential extensions, developing feature importance methods attributing the input features for the prediction.

\section{Acknowledgments}
This paper is based upon work supported by the U.S. Department of Energy, Office of Science, Office of Advanced Scientific Computing Research and Office of Biological and Environmental Research, Scientific Discovery through Advanced Computing (SciDAC) program under Award Number 9233218CNA000001.
This work is also partly sponsored by Science Undergraduate Laboratory Internship (SULI) at Brookhaven National Laboratory.

\bibliography{aaai24}


\end{document}